\documentclass[a4paper]{article}
\usepackage{multirow}
\usepackage{INTERSPEECH2020}
\usepackage{url}
\title{Multi-modal embeddings using multi-task learning for emotion recognition}
\name{Aparna Khare, Srinivas Parthasarathy, Shiva Sundaram}
\address{  Amazon.com, Sunnyvale, CA}
\email{apkhare,parsrini,sssundar@amazon.com}

\begin{document}

\maketitle
\begin{abstract}
 
  General embeddings like word2vec, GloVe and ELMo have shown a lot of success in natural language tasks. The embeddings are typically extracted from models that are built on general tasks such as skip-gram models and natural language generation. In this paper, we extend the work from natural language understanding to multi-modal architectures that use audio, visual and textual information for machine learning tasks. The embeddings in our network are extracted using the encoder of a  transformer model trained using multi-task training. We use person identification and automatic speech recognition as the tasks in our embedding generation framework. We tune and evaluate the embeddings on the downstream task of emotion recognition and demonstrate that on the CMU-MOSEI dataset, the embeddings can be used to improve over previous state of the art results. 
\end{abstract}

\section{Introduction}

Humans possess the ability to encode and express a wide range of intricate verbal and non-verbal cues based on goal and context. This has evolved into a complementary ability to detect nuanced cues in everyday communication.  This ability is a result of top-down processing \cite{devito2000human} where based on context and learning humans are able to encode and decode person to person information flow efficiently. Context is typically set by what is being communicated and how through multi-modal cues.  Inspired by this, several studies have shown that multi-modal input to the systems can improve accuracy on  tasks involving human communication, such as speech recognition \cite{afouras2018deep, dupont2000audio}, emotion recognition \cite{schuller2011avec} and speaker recognition \cite{sadjadi20202019}.

Recently, use of generalized feature representations have become prevalent in the computer vision and natural language research. Computer vision tasks like object detection and semantic segmentation show improved accuracy when the features from the images are extracted using models trained on large amounts of data like ImageNet \cite{girshick2014rich}.  In the natural learning literature, generalized embeddings like GloVe and word2vec have demonstrated state of the art performance in several tasks like word similarity, word analogy and named entity recognition\cite{pennington2014glove}.  For speech applications like automatic speech recognition (ASR), speaker recognition and paralinguistics it is still traditional to use hand-crafted features like MFCCs, LFBEs or features from toolkits like openSMILE \cite{schuller2011avec}.  However, it has also been demonstrated that features learned directly from audio can improve performance when the amount of training data is large enough \cite{palaz2015convolutional}.  

The research in the various domains has demonstrated that  transfer learning with models trained on large datasets can improve accuracy on subsequent tasks. This is especially important when the size of the labeled datasets is not large. There are a variety of multi-modal tasks like emotion recognition which still do not have large amounts of publicly available datasets.  Motivated by this, we propose a model to learn  embeddings that combine the features from audio, video, and text modalities to improve the performance on downstream tasks. The main contribution of this paper is to understand if we can leverage large datasets to build these representations that can outperform the models built for specific tasks where the datasets are limited. For our work, we use emotion recognition as the downstream task to evaluate the embeddings. In practical applications, it is possible that all modalities are not available to the machine learning system for inference. For example, for any applications that use video from web-based applications, any disturbance in the communication network can lead to missing audio or visual input. This leads to the second objective of our study; to perform ablation studies to understand the impact of the missing modality, and understand how to compensate for it.

This paper is organized as follows; in Section \ref{priorwork}, we discuss prior work in multi-modal tasks and embedding generation techniques.  Our proposed technique for embedding extraction is presented in Section \ref{sec:multitask}. In \ref{experiments} we discuss the training setup and data. Finally, we present our results in Section \ref{results} and conclude in Section \ref{discussion}.

\section{Prior work}
\label{priorwork}

Leveraging large datasets for representation learning, like billions of words from the Wikipedia dataset to train BERT or millions of images from ImageNet to train ResNET \cite{he2016deep}, has become quite popular in machine learning to improve performance on tasks with smaller labeled datasets. Tseng et al. demonstrate in \cite{tseng2019multimodal} that a model trained on the language modeling (LM) task, which was pre-trained on the 1 Billion Word Language Model Benchmark, can be used to extract ELMo-like sentence embeddings using audio and word tokens to produce state of the art results on the emotion recognition task.  Similarly, Tseng et al. also used an encoder-decoder architecture trained on a skip-though vector task on 30 million sentence pairs from the OpenSubtitles Corpus,  to extract text-only embeddings in \cite{tseng2019unsupervised} . They further show that in conjunction with weakly supervised learning, the features extracted from such a network can improve performance for sentiment analysis. In the natural language domain, the BERT model trained on a 3 billion words from the Wikipedia and BookCorpus corpora has demonstrated state of the art results on various natural language tasks like natural language inference, sentence classification, sentiment analysis and question answering \cite{devlin2018bert} . Although not widely adopted, there has been similar work in the speech domain; Chung et al. introduce speech2vec  in \cite{chung2018speech2vec} , which is trained similarly as word2vec but with speech as its input. This paper focuses on generating embeddings that have semantic information and semantically similar words that are close in the latent space. There has also been work done to generate embeddings from speech which are similar for similar sounding words \cite{el2019learning}. 

While models like BERT use unsupervised techniques using a single task to learn generalized representations, there has been a parallel area of work that shows how multi-task training can also help improve performance. As described above, \cite{tseng2019unsupervised} employs multi-task learning to improve performance on the emotion recognition task.  \cite{akhtar2019multi} also shows that multi-task training on sentiment and emotion recognition improves performance on both tasks. Luong et al. demonstrate  in \cite{luong2015multi}  that multi-task training with shared encoder and individual decoders for the machine translation and other supervised and unsupervised tasks like skip-though vector generation and image caption generation can improve performance on the translation task, thus helping the model learn more generalized representations.

An additional challenge that multi-modal systems present is combining multiple modalities in the system. Several papers have focused on late fusion; combining features from the different modalities after a few neural network layers \cite{tzirakis2017end}. \cite{tseng2019multimodal} employs a feature fusion at the frame level using a gated convolutional architecture. More recently, cross-modal transformers  have been employed in order to project one modality into another \cite{tsai2019multimodal}, which have been successfully applied to sentiment analysis \cite{tsai2019multimodal} and speech recognition \cite{paraskevopoulos2020multiresolution}. \cite{collell2017imagined} employs a neural network to extract multi-modal embeddings from visual features by applying a loss function that forces the embedding to be similar to the GloVe embedding of the entity in the image.  

In this paper, we extend the prior work in three main ways. Firstly, we combine the audio, visual and text modalities together in order to learn tri-modal representations from large datasets. Additionally, instead of using a single task for training the model, we use a multi-task training architecture to make the embeddings more general in nature. Finally, we employ the cross-modal transformer to train our model. Transformers lend themselves naturally to multi-modal sequences of different lengths as they allow for an easy combination of modalities without the need to specifically align them as shown in  \cite{tsai2019multimodal}.

\section{Multitask training for learning embeddings}
\label{sec:multitask}

\begin{figure}[t]
  \centering
  \includegraphics[width=\linewidth]{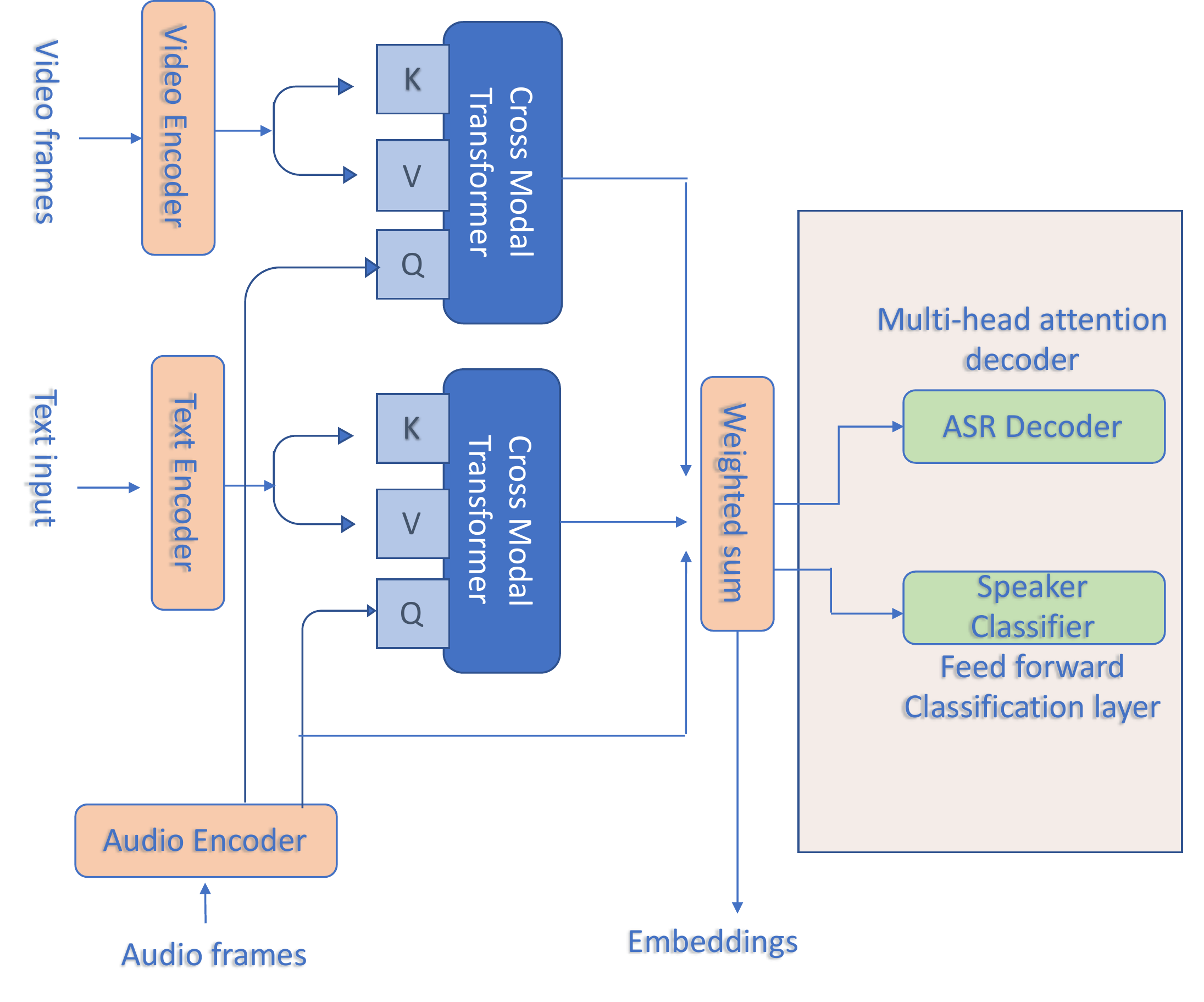}
  \caption{Transformer based multi-task architecture}
  \label{fig:multitask}
  \vspace{-14mm}
\end{figure}

We use two tasks for training our model; character level speech recognition and person identification.  ASR is a natural equivalent of the language modeling task for the audio domain, and the language modeling task has been applied with success for learning generalized embeddings.  The choice of the person identification task is mainly due to the availability of open source datasets with labels for the task.  We employ the architecture proposed in \cite{paraskevopoulos2020multiresolution} in order to project the features in the text and visual domain to the audio domain shown in Figure \ref{fig:multitask}. This is an alternative to the traditional early and late fusion techniques for combining features from different modalities; the cross-modal attention module computes the attention weights between features from two modalities, and weighs the features that are correlated higher, thus paying more attention to the features that are more important for the task. There is a joint encoder that comprises of an audio only encoder, and 2 cross-modal transformer components which project the video and text features to the audio space. The final sequential output from the encoder is a simple weighted sum of the various modalities.  For the ASR decoder, we use the Transformer decoder, whereas for the person identification task we average the embeddings over a given sequence and use an affine transformation layer for classifying the speaker.  The final loss function is a weighted sum of the losses over the 2 losses. In our setup, we use a weight of 0.2 for the person identification task and 0.8 for the ASR task. This quantity was not tuned and we will work on tuning and observing the effects of the weights in our future work.  The ASR task is more general in nature; since it involves learning the language which inherently has semantic meaning, we intuitively choose a higher weight on this task.

For our model, we choose a weight of 0.4 for the audio and visual modalities and 0.2 for the text modality. The text  input to the system in the form of GloVe embeddings can be reverse mapped to the characters, therefore we heuristically chose a lower weight on the text modality in order for the model to focus more learning on the audio and video modalities. The text input can be considered the equivalent of a thought vector in this case and is still relevant to the ASR task. Similarly, for the person identification task, text is an auxiliary input which can be used to make the system more text dependent.  For the cross-modal transformer component, we chose to transform the text and visual domains into the audio domain as explained in \cite{tsai2019multimodal}, but we could have chosen to map the audio and textual domains to visual modality as well. In our future work, we will evaluate the impact of this choice.

Other tasks like skip-thought vector and masked language modeling task, that have been used in the natural language domain can also be applied to this kind of multi-task architecture. These tasks, however, present certain challenges when the audio and visual modalities are involved. For the masked LM task, the model tries to predict a word that has been masked in the input. For audio and visual modalities, this would involve aligning the text with the audio  to obtain the time alignment for each word, and computing out the corresponding masks. The skip-thought vector task would require the dataset to have continuity. This continuity is easily provided for datasets from the text domain, where the data is typically obtained from documents or Wikipedia. When using videos, preparing such a dataset would require finding sentence boundaries, and segmenting the videos in order to find subsequent sentences to set up the task. These alignments and segmentation tasks can be accomplished by using an ASR system, but introduce complexity in the process. We will include these tasks in our future work.

\begin{table*}[tp]
\begin{center}
\caption{Baseline results on the CMU-MOSEI dataset. TF is the transformer-based model}
\vspace{-4mm}
\label{tab:moseibaseline}
\resizebox{\textwidth}{!}{\begin{tabular}{c c c c c c c c c c c c c c c}
\toprule
\multirow{2}{*}{\parbox{0\linewidth}}{\bf Model} &\multicolumn{2}{c}{\bf Happy} &\multicolumn{2}{c}{\bf Sad} &\multicolumn{2}{c}{\bf Anger} &\multicolumn{2}{c}{\bf{Surprise}} &\multicolumn{2}{c}{\bf{Disgust}} &\multicolumn{2}{c}{\bf {Fear}}&\multicolumn{2}{c}{\bf {Average}}\\ 
\multirow{2}{*}{}& \multicolumn{6}{c}{}\\
 & WA & F1 & WA  & F1 & WA & F1& WA & F1 & WA & F1 & WA & F1 & WA & F1 \\ 
\midrule
M-ELMo + NN \cite{tseng2019multimodal}(A+T) & 67.0	&65.2	&63.1	&72.0	&65.8	&\bf{74.7}&	\bf{63.8}	&83.3	&74.2&81.7	& 63.2 &85.1	&66.2	&77.0 \\
\midrule
Graph-MFN \cite{liang2018computational} & 66.3 & 66.3 & 60.4 & 66.9 &  62.6 & 72.8 & 53.7 & 85.5 &  69.1 & 76.6 & 62.0  & \bf{89.9} &62.3 &76.3\\
 \midrule 
CIM-Att-STL  \cite{akhtar2019multi} & 53.6 & 67.0 & 61.4 & 72.4 &\bf{66.8}& 75.9& 60.6 & 86.0 & 72.7 & 81.9 & 62.2 & 87.9 & 62.8 & \bf{78.6} \\
\midrule 
Late Fusion  & 66.2	 & 66.3	& 63.7 & 	71.8& 	66.2& 	73.6& 	62.2& 	87.8& 	74.1& \bf{83.3}& 	62.3& 	86.6	& 65.8	&  78.2\\
\midrule
TF  & \bf{67.8} & \bf{67.5} &\bf{63.9}& \bf{72.9} &\bf{66.8}& 74.5&  60.6 &\bf{88.0} & \bf{74.8}& 82.3 &\bf{64.4}& 86.1&\bf{66.4} & 78.5 \\
\bottomrule
\end{tabular}}
\end{center}
\vspace{-4mm}
\end{table*}

\begin{table*}[tp]
\begin{center}
\caption{Weighted accuracy and F1-score with embeddings learned on the VoxCeleb2 dataset}
\vspace{-4mm}
\label{tab:moseiresults}
\resizebox{\textwidth}{!}{\begin{tabular}{c c c c c c c c c c c c c c c}
\toprule
\multirow{2}{*}{\parbox{0\linewidth}}{\bf Model} &\multicolumn{2}{c}{\bf Happy} &\multicolumn{2}{c}{\bf Sad} &\multicolumn{2}{c}{\bf Anger} &\multicolumn{2}{c}{\bf{Surprise}} &\multicolumn{2}{c}{\bf{Disgust}} &\multicolumn{2}{c}{\bf {Fear}}&\multicolumn{2}{c}{\bf {Average}}\\ 
\multirow{2}{*}{}& \multicolumn{6}{c}{}\\
 & WA & F1 & WA  & F1 & WA & F1& WA & F1 & WA & F1 & WA & F1 & WA & F1 \\ 
\midrule
Baseline  & \bf{67.8} &  67.5 &63.9& \bf{72.9} &66.8& 74.5&  60.6 & \bf{88.0} &74.8& 82.3 &\bf{64.4}& 86.1&66.4&78.5 \\
\midrule
Embeddings from the Multi-task model &  67.5 & \bf{67.7} & \bf{65.4} &71.3 &\bf{67.0} & \bf{75.8}  &\bf{63.9} &87.8 & \bf{74.9} &\bf{83.4} &64.2&\bf{ 86.9} &\bf{67.1} &\bf{78.8} \\
\bottomrule
\end{tabular}}
\end{center}
\vspace{-4mm}
\end{table*}

\begin{table*}[ht]
\caption{Ablation studies with and without learned embeddings. The metrics in the table are average weighted accuracy and F1-scores over the 6 emotions.}
\vspace{-5mm}
\label{tab:ablation}
\begin{center}
\begin{tabular}{c c c c c c c c c}
\toprule
\multirow{2}{*}{\parbox{0\linewidth}}{\bf Model} &\multicolumn{2}{c}{\bf A} &\multicolumn{2}{c}{\bf{A+V}} &\multicolumn{2}{c}{\bf{A+T}} &\multicolumn{2}{c}{\bf {A+V+T}}\\ 
\multirow{2}{*}{}& \multicolumn{6}{c}{}\\
 & WA & F1 & WA  & F1 & WA & F1& WA & F1 \\ 
 \midrule
 Baseline & 57.2& 74.4&  60.3 & 76.3 & 65.8& 77.8 & 66.5 & 78.5\\
\midrule
Embeddings from the Multi-task model & 57.6 & 75.1 & 60.9 & 76.4 & 66.2 & 78.4 & 67.1 & 78.8	 \\

\bottomrule
\end{tabular}
\end{center}
\vspace{-8mm}
\end{table*}

\section{Experimental setup}
\label{experiments}

\subsection{Datasets and training setup}
For training the multi-task transformer models, we use the VoxCeleb2 dataset \cite{Chung18b}. This train partition of the dataset has 1.1 million examples. Since there are no transcriptions available with this dataset, we use a TDNN ASR model trained with Kaldi on Librispeech dataset in order to extract text labels for this dataset \cite{peddinti2015time}. We use the standard recipe available in Kaldi in order to train the models.  We extract 40-dimensional Log-Filter bank energy (LFBE) features using a 10ms frame rate to represent the audio, 4096-dimensional features extracted from the VGG-16 model to represent the visual modality and 300-dimensional GloVe vector to represent the text modality. The model is trained using pytorch with the learning schedule described in \cite{dong2018speech} We stack 5-frames of the LFBE features for a final feature audio dimensionality of 200. The first operation in the transformer encoder is a dot-product attention layer, which computes the attention weights between the query $Q$ and the key $K$ as softmax$(QK^T/\sqrt(d))$, where $d$ is the dimensionality of $K$ (or $Q$), and they are matrices of dimension $n \times d$ where $n$ is the sequence length. By stacking the features to reduce the sequence length, we reduce the memory required to compute the attention maps. Since the Voxceleb2 dataset consists of data from multiple languages, we first filter out the data based on the likelihood from the ASR decoding to pick only the training examples in English. For all our experiments, we use only the dev portion of the Voxceleb2 dataset.  After filtering, we have 1 million utterances for training with 5994 unique speakers.  We use the architecture described in Section \ref{sec:multitask} to train the model. The model  has a dimensionality of 512 with 4 encoder layers for audio and cross-modal encoders, with 2 decoder layers and 4 attention heads, and a feed forward layer of dimension 200 for the transformer.  The cross-modal encoder has 4 encoder layers with 4 attention heads. This model architecture was not optimized.
We evaluate our embeddings on the downstream task of emotion recognition. We choose the CMU-MOSEI dataset for this purpose since it has all the modalities available \cite{liang2018computational}. The dataset consists of 6 emotions; happy, sad, angry, disgust, surprise and fear. It  contains 23,453 single-speaker video segments from YouTube which have been manually transcribed and annotated for sentiment and emotion. Each segment can have multiple emotions. The labels for each class are on a Likert scale of $[0,3]$; we binarize the problem by giving a segment a label of 1 if it has a Likert score greater than 0. The classes thus obtained represent presence versus no presence of a specific emotion.
For the baseline model, we tried two different approaches to get the best possible baseline. The first model is a simple late fusion model trained with the same feature inputs as the transformer model. We use a 2-layer bidirectional-GRU model with 200 nodes for each modality. The output from the last time step from each modality is then concatenated and a linear layer of size 100 is used for classification. We train a 6-class classifier using binary cross entropy loss over each class and use weighted training in order to compensate for the imbalanced nature of the dataset.  The second architecture we tried is the same cross-modal architecture as show in Figure \ref{fig:multitask}, where the output of the encoder layer goes to a linear classification layer.  To train the transformer model, we use equal weights on all modalities. We experimented with learned weights but the final weights learned by the model were not very different from equal weights so we choose to use this as a fixed parameter.  

In order to leverage the learned embeddings from the VoxCeleb dataset, we use the model trained with multi-task training as a pre-trained model and tune it with the CMU-MOSEI dataset on the emotion recognition task. We chose to tune the embeddings and not fix them based on prior research from the natural language domain where researchers have demonstrated that tuning the word2vec embeddings for the language modeling task can improve performance. \cite{komiya2018investigating}.
\vspace{-2mm}
\section{Results}
\label{results}

For all our evaluations, we report the weighted accuracy (WA) and F1-score for each emotion and the average metrics over the 6 emotions. The threshold for the both metrics was optimized separately on the development set.  For all the reported results on the emotion recognition task, the model is randomly initialized and trained  10 different times. The best model is chosen based on the average of the weighted accuracy  and F1-scores over the dev set over the 10 runs, as suggested in \cite{tseng2019multimodal}.  

\subsection{Baseline results}

Table \ref{tab:moseibaseline} shows the results of the two baseline models. We have also included results from other publications in order to compare our metrics. From \cite{akhtar2019multi}, we use the multi-task model since it demonstrated the best results. As the results demonstrate, the transformer-based baseline outperforms both the reported results in literature as well as the late fusion baseline model that we trained for most of the emotion categories in the dataset on majority of the emotions, as well as on the average weighted accuracy over the emotions. This result is not unexpected, since Tseng et al.  already demonstrated in \cite{tsai2019multimodal} that this architecture outperforms the late fusion architecture for the MOSEI sentiment task. They however did not report results on the emotion task for the dataset.  For the remainder of the paper, our baseline will refer to the transformer-based baseline in this table. Note that the external publications used a different set of features for the audio and visual modalities training their models, which is why we present the late fusion results with our feature set.

\subsection{Results with embeddings}
Table \ref{tab:moseiresults} shows the results of the experiment where we use the model trained with multi-task training on the VoxCeleb2 dataset to initialize the transformer model for the emotion recognition task. The model outperforms the baseline by more than 1\% absolute for the weighted accuracy for the sad and surprise emotions, and F1-score for the anger emotion. We observe a degradation in the weighted accuracy of the fear emotion. The rest of the metrics are comparable to the baseline model. The improvements on the weighted accuracy of the surprise and sad emotions are significant as the 95\% confidence intervals are non-overlapping.

To analyze the results obtained from the embeddings, we look at the distribution of emotions in the VoxCeleb2 dataset. The distribution on the development set is not available, however the authors of the dataset have created a subset of the dataset called EmoVoxceleb \cite{Albanie18a}. This dataset is a subset containing 100k videos from the Voxceleb2 dataset. Since the authors haven't mentioned a sampling strategy, we assume that this data was randomly sampled. In \cite{Albanie18a}, Albanie et al. present a frame level distribution of the emotions in the dataset. Neutral, happy, surprise and sadness are the dominant emotions in the dataset, whereas fear and disgust are the least frequent. Under the assumption that the Voxceleb2 dataset has a similar distribution of emotions, we posit that the improvements on the sad and surprise emotions and degradation in fear emotion can be attributed to this distribution. 

\subsection{Ablation studies}

As mentioned in the introduction, for practical applications we would like the understand the impact of missing input modalities. To understand this, the first experiment we perform is to remove the video and/or text inputs to both the baseline as well as the model pre-trained model during inference only.  Since in our architecture we use the cross-modal attention to transform the video and text modalities to the audio domain, the architecture doesn't allow for a missing audio modality. In order to run the ablation study, we just use a 0 weight on the missing modality during the weighted combination of the outputs of the audio encoder and the cross-modal transformers in the model to compute the final embedding. 

Table \ref{tab:ablation} show the results with different combination of the modalities.  The results show that in the case of missing modalities, the embeddings improve over the baseline even more than when all the modalities are present. When both the video and text modalities are absent, the model performs the worst. With text and audio input, the performance of the baseline and embeddings-based model is close to the performance with no ablation. Since the GloVe embeddings, which are already pre-trained to represent semantic meaning with over 6 billion tokens \cite{pennington2014glove}, they already provide meaningful features to the system in order to get good performance. Prior work \cite{tsai2019multimodal,majumder2018multimodal} has shown that for uni-modal emotion recognition, text  based models outperforms audio and visual based models, and our ablation results demonstrate the same; adding text to the input audio boosts the weighted average by 8.6\% absolute as compared to adding visual information that improves it by 3.3\%.

\section{Conclusions and Future work}
\label{discussion}
In this paper, we present state of the art results on the CMU-MOSEI emotion recognition task using a cross-modal attention based transformer architecture. We demonstrate that by using a multi-task architecture, we can leverage a large multi-modal dataset like VoxCeleb in order to learn general embeddings that can improve the performance on a downstream task like emotion recognition.  

Our future work will address one of the major drawbacks of our architecture in its dependency on audio which doesn't allow for missing audio during inference. In addition, for text input to the system, we require either a human or a machine based transcription. Prior work, however, has demonstrated that we can learn embeddings from audio that capture the semantic meaning \cite{chung2018speech2vec}. By incorporating this in the model, we should be able to eliminate the need for the text input altogether.  Both the CMU-MOSEI and Voxceleb2 datasets are obtained from YouTube but there could be some mismatch due to the way the videos were selected. This could be bridged by adapting the embeddings on the CMU-MOSEI dataset for improved performance. We would like to increase the number of tasks in our multi-task architecture in order to make the representations even more generalized. Specifically, experimenting with visual tasks like landmark detection which would provide useful information for tasks involving affect recognition. An additional future direction is to try unsupervised methods like skip-thought vector or other self-supervised training methods. Another interesting approach to consider is to combine the multi-task learning with  weak supervision as suggested in \cite{tseng2019unsupervised} for each of the tasks in question in order to make the embeddings from discriminative. Finally, we will evaluate our technique on more downstream tasks to study what tasks the representations generalize well to. 

\bibliographystyle{IEEEtran}

\bibliography{mybib}

\begin{thebibliography}{10}
\providecommand{\url}[1]{#1}
\csname url@samestyle\endcsname
\providecommand{\newblock}{\relax}
\providecommand{\bibinfo}[2]{#2}
\providecommand{\BIBentrySTDinterwordspacing}{\spaceskip=0pt\relax}
\providecommand{\BIBentryALTinterwordstretchfactor}{4}
\providecommand{\BIBentryALTinterwordspacing}{\spaceskip=\fontdimen2\font plus
\BIBentryALTinterwordstretchfactor\fontdimen3\font minus
  \fontdimen4\font\relax}
\providecommand{\BIBforeignlanguage}[2]{{%
\expandafter\ifx\csname l@#1\endcsname\relax
\typeout{** WARNING: IEEEtran.bst: No hyphenation pattern has been}%
\typeout{** loaded for the language `#1'. Using the pattern for}%
\typeout{** the default language instead.}%
\else
\language=\csname l@#1\endcsname
\fi
#2}}
\providecommand{\BIBdecl}{\relax}
\BIBdecl

\bibitem{devito2000human}
J.~A. DeVito, S.~O'Rourke, and L.~O'Neill, \emph{Human communication}.\hskip
  1em plus 0.5em minus 0.4em\relax Longman, 2000.

\bibitem{afouras2018deep}
T.~Afouras, J.~S. Chung, A.~Senior, O.~Vinyals, and A.~Zisserman, ``Deep
  audio-visual speech recognition,'' \emph{IEEE transactions on pattern
  analysis and machine intelligence}, 2018.

\bibitem{dupont2000audio}
S.~Dupont and J.~Luettin, ``Audio-visual speech modeling for continuous speech
  recognition,'' \emph{IEEE transactions on multimedia}, vol.~2, no.~3, pp.
  141--151, 2000.

\bibitem{schuller2011avec}
B.~Schuller, M.~Valstar, F.~Eyben, G.~McKeown, R.~Cowie, and M.~Pantic, ``Avec
  2011--the first international audio/visual emotion challenge,'' in
  \emph{International Conference on Affective Computing and Intelligent
  Interaction}.\hskip 1em plus 0.5em minus 0.4em\relax Springer, 2011, pp.
  415--424.

\bibitem{sadjadi20202019}
S.~O. Sadjadi, C.~S. Greenberg, E.~Singer, D.~A. Reynolds, L.~Mason, and
  J.~Hernandez-Cordero, ``The 2019 nist audio-visual speaker recognition
  evaluation,'' \emph{Proc. Speaker Odyssey (submitted), Tokyo, Japan}, 2020.

\bibitem{girshick2014rich}
R.~Girshick, J.~Donahue, T.~Darrell, and J.~Malik, ``Rich feature hierarchies
  for accurate object detection and semantic segmentation,'' in
  \emph{Proceedings of the IEEE conference on computer vision and pattern
  recognition}, 2014, pp. 580--587.

\bibitem{pennington2014glove}
\BIBentryALTinterwordspacing
J.~Pennington, R.~Socher, and C.~D. Manning, ``Glove: Global vectors for word
  representation,'' in \emph{Empirical Methods in Natural Language Processing
  (EMNLP)}, 2014, pp. 1532--1543. [Online]. Available:
  \url{http://www.aclweb.org/anthology/D14-1162}
\BIBentrySTDinterwordspacing

\bibitem{palaz2015convolutional}
D.~Palaz, M.~M. Doss, and R.~Collobert, ``Convolutional neural networks-based
  continuous speech recognition using raw speech signal,'' in \emph{2015 IEEE
  International Conference on Acoustics, Speech and Signal Processing
  (ICASSP)}.\hskip 1em plus 0.5em minus 0.4em\relax IEEE, 2015, pp. 4295--4299.

\bibitem{he2016deep}
K.~He, X.~Zhang, S.~Ren, and J.~Sun, ``Deep residual learning for image
  recognition,'' in \emph{Proceedings of the IEEE conference on computer vision
  and pattern recognition}, 2016, pp. 770--778.

\bibitem{tseng2019multimodal}
S.-Y. Tseng, P.~Georgiou, and S.~Narayanan, ``Multimodal embeddings from
  language models,'' \emph{arXiv preprint arXiv:1909.04302}, 2019.

\bibitem{tseng2019unsupervised}
S.-Y. Tseng, B.~Baucom, and P.~Georgiou, ``Unsupervised online multitask
  learning of behavioral sentence embeddings,'' \emph{PeerJ Computer Science},
  vol.~5, p. e200, 2019.

\bibitem{devlin2018bert}
J.~Devlin, M.-W. Chang, K.~Lee, and K.~Toutanova, ``Bert: Pre-training of deep
  bidirectional transformers for language understanding,'' \emph{arXiv preprint
  arXiv:1810.04805}, 2018.

\bibitem{chung2018speech2vec}
Y.-A. Chung and J.~Glass, ``Speech2vec: A sequence-to-sequence framework for
  learning word embeddings from speech,'' \emph{Proc. Interspeech 2018}, pp.
  811--815, 2018.

\bibitem{el2019learning}
M.~El-Geish, ``Learning joint acoustic-phonetic word embeddings,'' \emph{arXiv
  preprint arXiv:1908.00493}, 2019.

\bibitem{akhtar2019multi}
M.~S. Akhtar, D.~S. Chauhan, D.~Ghosal, S.~Poria, A.~Ekbal, and
  P.~Bhattacharyya, ``Multi-task learning for multi-modal emotion recognition
  and sentiment analysis,'' \emph{arXiv preprint arXiv:1905.05812}, 2019.

\bibitem{luong2015multi}
M.-T. Luong, Q.~V. Le, I.~Sutskever, O.~Vinyals, and L.~Kaiser, ``Multi-task
  sequence to sequence learning,'' \emph{arXiv preprint arXiv:1511.06114},
  2015.

\bibitem{tzirakis2017end}
P.~Tzirakis, G.~Trigeorgis, M.~A. Nicolaou, B.~W. Schuller, and S.~Zafeiriou,
  ``End-to-end multimodal emotion recognition using deep neural networks,''
  \emph{IEEE Journal of Selected Topics in Signal Processing}, vol.~11, no.~8,
  pp. 1301--1309, 2017.

\bibitem{tsai2019multimodal}
Y.-H.~H. Tsai, S.~Bai, P.~P. Liang, J.~Z. Kolter, L.-P. Morency, and
  R.~Salakhutdinov, ``Multimodal transformer for unaligned multimodal language
  sequences,'' pp. 6558--6569, 2019.

\bibitem{paraskevopoulos2020multiresolution}
G.~Paraskevopoulos, S.~Parthasarathy, A.~Khare, and S.~Sundaram, ``Multimodal
  and multiresolution speech recognition with transformers,'' in
  \emph{Proceedings of the 58th Annual Meeting of the Association for
  Computational Linguistics}, 2020, pp. 2381--2387.

\bibitem{collell2017imagined}
G.~Collell, T.~Zhang, and M.-F. Moens, ``Imagined visual representations as
  multimodal embeddings,'' in \emph{Thirty-First AAAI Conference on Artificial
  Intelligence}, 2017.

\bibitem{liang2018computational}
P.~Liang, R.~Salakhutdinov, and L.-P. Morency, ``Computational modeling of
  human multimodal language: The mosei dataset and interpretable dynamic
  fusion,'' 2018.

\bibitem{Chung18b}
J.~S. Chung, A.~Nagrani, and A.~Zisserman, ``Voxceleb2: Deep speaker
  recognition,'' in \emph{INTERSPEECH}, 2018.

\bibitem{peddinti2015time}
V.~Peddinti, D.~Povey, and S.~Khudanpur, ``A time delay neural network
  architecture for efficient modeling of long temporal contexts,'' in
  \emph{Sixteenth Annual Conference of the International Speech Communication
  Association}, 2015.

\bibitem{dong2018speech}
L.~Dong, S.~Xu, and B.~Xu, ``Speech-transformer: a no-recurrence
  sequence-to-sequence model for speech recognition,'' in \emph{2018 IEEE
  International Conference on Acoustics, Speech and Signal Processing
  (ICASSP)}.\hskip 1em plus 0.5em minus 0.4em\relax IEEE, 2018, pp. 5884--5888.

\bibitem{komiya2018investigating}
K.~Komiya and H.~Shinnou, ``Investigating effective parameters for fine-tuning
  of word embeddings using only a small corpus,'' in \emph{Proceedings of the
  Workshop on Deep Learning Approaches for Low-Resource NLP}, 2018, pp. 60--67.

\bibitem{Albanie18a}
S.~Albanie, A.~Nagrani, A.~Vedaldi, and A.~Zisserman, ``Emotion recognition in
  speech using cross-modal transfer in the wild,'' in \emph{ACM Multimedia},
  2018.

\bibitem{majumder2018multimodal}
N.~Majumder, D.~Hazarika, A.~Gelbukh, E.~Cambria, and S.~Poria, ``Multimodal
  sentiment analysis using hierarchical fusion with context modeling,''
  \emph{Knowledge-based systems}, vol. 161, pp. 124--133, 2018.

\end{thebibliography}

\end{document}